\documentclass{llncs}

\usepackage{graphicx}
\usepackage{enumitem}

\title{Machine learning for constraint solver design}
\subtitle{A case study for the \texttt{alldifferent} constraint}

\author{Ian Gent \and Lars Kotthof\/f \and Ian Miguel \and Peter Nightingale
\email{\{ipg,larsko,ianm,pn\}@cs.st-andrews.ac.uk}}
\institute{University of St Andrews}

\begin{document}

\maketitle

\begin{abstract}
Constraint solvers are complex pieces of software which require many design
decisions to be made by the implementer based on limited information. These
decisions affect the performance of the finished solver
significantly~\cite{survey}. Once a design decision has been made, it cannot
easily be reversed, although a different decision may be more appropriate for a
particular problem.

We investigate using machine learning to make these decisions automatically
depending on the problem to solve. We use the alldifferent constraint as a case
study. Our system is capable of making non-trivial, multi-level decisions that
improve over always making a default choice and can be implemented as part of a
general-purpose constraint solver.
\end{abstract}

\section{Introduction}

Constraints are a natural, powerful means of representing and reasoning about
combinatorial problems that impact all of our lives. Constraint solving is
applied successfully in a wide variety of disciplines such as aviation,
industrial design, banking, combinatorics and the chemical and steel industries,
to name but a few examples.

A \emph{constraint satisfaction problem}
(CSP~\cite{constraint-processing-dechter}) is a set of decision variables, each
with an associated domain of potential values, and a set of constraints.  An
assignment maps a variable to a value from its domain.  Each constraint
specifies allowed combinations of assignments of values to a subset of the
variables. A \emph{solution} to a CSP is an assignment to all the variables that
satisfies all the constraints. Solutions are typically found for CSPs through
systematic search of possible assignments to variables. During search,
constraint \emph{propagation} algorithms are used. These propagators make
inferences, usually recorded as domain reductions, based on the domains  of the
variables constrained and the assignments that satisfy the constraints.  If at
any point these inferences result in any variable having an empty domain then
search backtracks and a new branch is considered.

When implementing constraint solvers and modelling constraint problems, many
design decision have to be made -- for example what level of consistency to
enforce and what data structures to use to enable the solver to backtrack.
These decisions have so far been made mostly manually. Making the ``right''
decision often depends on the experience of the person making it.

We approach this problem using machine learning. Given a particular problem
class or problem instance, we want to decide \emph{automatically} which design
decisions to make.  This improves over the current state of the art in two ways.
First, we do not require humans to make a decision based on their experience and
data available at that time. Second, we can change design decisions for
particular problems.

Our system does not only improve the performance of constraint solving, but also
makes it easier to apply constraint programming to domain-specific problems,
especially for people with little or no experience in constraint programming.
It represents a significant step towards Puget's ``model and run''
paradigm~\cite{modelrun}.

We demonstrate that we can approach machine learning as a ``black box'' and use
generic techniques to increase the performance of the learned classifiers. The
result is a system which is able to dynamically decide which implementation to
use by looking at an unknown problem. The decision made is in general better
than simply relying on a default choice and enables us to solve constraint
problems faster.

\section{Background}

We are addressing an instance of the Algorithm Selection Problem~\cite{rice},
which, given variable performance among a set of algorithms, is to choose the
best candidate for a particular problem instance. Machine learning is an
established method of addressing this problem~\cite{lagoudakis,nudelman}.
Particularly relevant to our work are the machine learning approaches that have
been taken to configure, to select among, and to tune the parameters of solvers
in the related fields of mathematical programming, propositional satisfiability
(SAT), and constraints.

{\sc Multi-tac}~\cite{DBLP:journals/constraints/Minton96} configures a
constraint solver for a particular instance distribution. It makes informed
choices about aspects of the solver such as the search heuristic and the level
of constraint propagation. The Adaptive Constraint
Engine~\cite{DBLP:conf/cp/EpsteinFWMS02} learns search
heuristics from training instances.
SATenstein~\cite{DBLP:conf/ijcai/KhudaBukhshXHL09} configures stochastic local
search solvers for solving SAT problems.

An algorithm {\em portfolio} consists of a collection of algorithms, which can
be selected and applied in parallel to an instance, or in some (possibly
truncated) sequence. This approach has recently been used with great success in
SATzilla~\cite{DBLP:journals/jair/XuHHL08} and CP Hydra~\cite{cphydra}. In
earlier work Borrett {\em et al}~\cite{DBLP:conf/ecai/BorrettTW96} employed a
sequential portfolio of constraint
solvers. Guerri and Milano~\cite{guerri-milano-model-selection} use a
decision-tree based technique to select among a portfolio of constraint- and
integer-programming based solution methods for the bid evaluation problem.
Similarly, Gent {\em et al}~\cite{lazyecai} investigate decision trees to choose
whether to use lazy constraint learning~\cite{lazylearning} or not.

Rather than select among a number of algorithms, it is also possible to learn
parameter settings for a particular algorithm. Hutter {\em et
al}~\cite{DBLP:conf/cp/HutterHHL06} apply this method to local search. Ansotegui
{\em et al}~\cite{DBLP:conf/cp/AnsoteguiST09} employ a genetic algorithm to
tune the parameters of both local and systematic SAT solvers.

The \emph{alldifferent} constraint requires all variables which it is imposed on
to be pairwise alldifferent. For example alldiff($x_1,x_2,x_3$) enforces
$x_1\neq x_2$, $x_1\neq x_3$ and $x_2\neq x_3$.

There are many different ways to implement the alldifferent constraint. The
na\"ive version decomposes the constraint and enforces disequality on each pair
of variables. More sophisticated versions (e.g.~\cite{reginalldiff}) consider
the constraint as a whole and are able to do more propagation. For example an
alldifferent constraint which involves four variables with the same three
possible values each cannot be satisfied, but this knowledge cannot be derived
when just considering the decomposition into pairs of variables. Further
variants are discussed in~\cite{hoeve_alldifferent_2001}.

Even when the high-level decision of how much propagation to do has been made,
a low-level decision has to be made on how to implement the constraint. For an
in-depth survey of the decisions involved, see~\cite{petealldiff}.

We make both decisions and therefore combine the selection of an algorithm (the
na\"ive implementation or the more sophisticated one) and the tuning of
algorithm parameters (which one of the more sophisticated implementations to
use). Note that we restrict the implementations to the ones that the Minion
constraint solver~\cite{minion} provides. In particular, it does not provide a
bounds consistency propagator.

\section{The benchmark instances and solvers}

We evaluated the performance of the different versions of the alldifferent
constraint on two different sets of problem instances. The first one was used
for learning classifiers, the second one only for the evaluation of the learned
classifiers.

The set we used for machine learning consisted of 277 benchmark instances from
14 different problem classes. It has been chosen to include as many instances as
possible whatever our expectation of which version of the alldifferent
constraint will perform best.

The set to evaluate the learned classifiers consisted of 1036 instances from 2
different problem classes that were not present in the set we used for machine
learning. We chose this set for evaluation because the low number of different
problem classes makes it unsuitable for training.

Our sources are Lecoutre's XCSP repository~\cite{xcsprepo} and our own stock of
CSP instances.  The reference constraint solver used is Minion~\cite{minion}
version 0.9 and its default implementation of the alldifferent constraint
\texttt{gacalldiff}. The experiments were run with binaries compiled with g++
version 4.4.3 and Boost version 1.40.0 on machines with 8 core Intel E5430
2.66GHz, 8GB RAM running CentOS with Linux kernel 2.6.18-164.6.1.el5 64Bit.

We imposed a time limit of 3600 seconds for each instance. The total number of
instances that no solver could solve solve because of a time out was 66 for the
first set and 26 for the second set. We took the median CPU time of 3 runs for
each problem instance.

\begin{figure}
\begin{center}
\includegraphics[width=.8\textwidth]{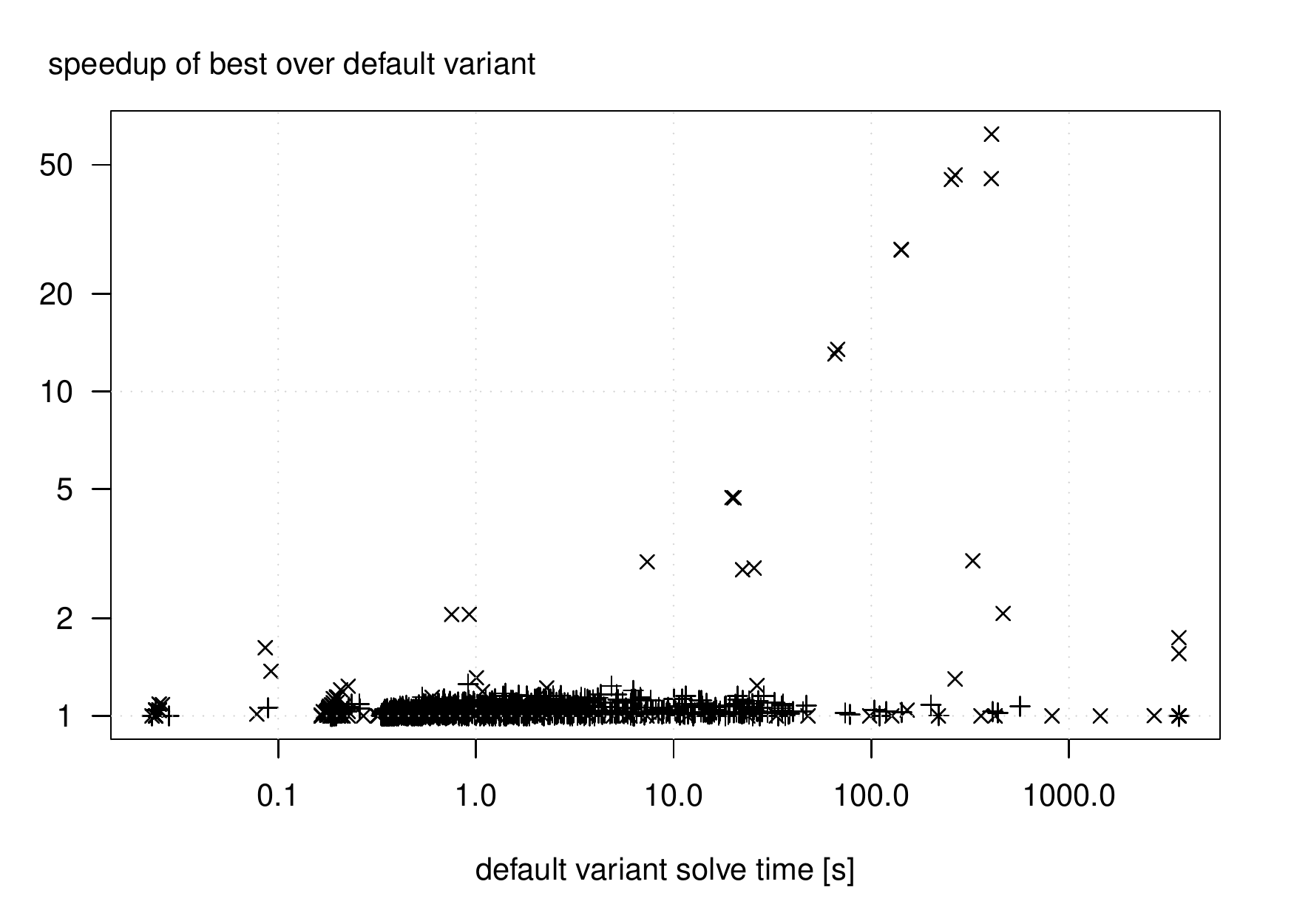}
\end{center}
\vspace*{-3em}
\caption{Potential speedup a decision algorithm could achieve over always making
the default decision. The crosses represent the instances of the first data
set, the pluses the instances of the second data set. A speedup of one
means that the default version of alldifferent is the fastest version, a speedup
of two means that the fastest version of alldifferent is twice as fast as the
default version.}
\label{fig:speedup-potential}
\end{figure}

As Figure~\ref{fig:speedup-potential} shows, adapting the implementation
decision to the problem instead of always choosing a standard implementation has
the potential of achieving significant speedups on some instances of the first
set of benchmark instances and speedups of up to 1.2 on the second set.

We ran the problems with 9 different versions of the alldifferent
constraint -- the na\"ive version which is operationally equivalent to the
binary decomposition and 8 different implementations of the more sophisticated
version which achieves generalised arc consistency (see~\cite{petealldiff}). The
amount of search done by the 8 versions which implement the more sophisticated
algorithm was the same. The variables and values were searched in the order they
were specified in in the model of the problem instance.

The instances, the binaries to run them, and everything else required to
reproduce our results are available on request.

\section{Instance attributes and their measurement}\label{sec:attrs}


We measured 37 attributes of the problem instances. They describe a wide range
of features such as constraint and variable statistics and a number of
attributes based on the primal graph. The primal graph $g=\langle V,E\rangle$
has a vertex for every CSP variable, and two vertices are connected by an edge
iff the two variables are in the scope of a constraint together.

\begin{description}
\item[Edge density] The number of edges in $g$ divided by the number of pairs of
distinct vertices.

\item[Clustering coefficient] For a vertex $v$, the set of neighbours of $v$ is
$n(v)$. The edge density among the vertices $n(v)$ is calculated. The
clustering coefficient is the mean average of this local edge density for all
$v$~\cite{watts-strogatz-small-world}
. It is intended to be a measure of the
local cliqueness of the graph. This attribute has been used with machine
learning for a model selection problem in constraint
programming~\cite{guerri-milano-model-selection}.

\item[Normalised degree] The normalised degree of a vertex is its degree
divided by $|V|$. The minimum, maximum, mean and median normalised degree are
used.

\item[Normalised standard deviation of degree] The standard deviation of
vertex degree is normalised by dividing by $|V|$.

\item[Width of ordering] Each of our benchmark instances has an
associated variable ordering. The width of a vertex $v$ in an ordered graph is
its number of \textit{parents} (i.e.\ neighbours that precede $v$ in the
ordering). The width of the ordering is the maximum width over all
vertices~\cite{constraint-processing-dechter}. The width of the ordering
normalised by the number of vertices was used.

\item[Width of graph] The width of a graph is the minimum width over all
possible orderings. This can be calculated in polynomial
time~\cite{constraint-processing-dechter}, and is related to some tractability
results. The width of the graph normalised by the number of vertices was used.

\item[Variable domains] The quartiles and the mean value over the domains of all
variables.

\item[Constraint arity] The quartiles and the mean of the arity of all
constraints (the number of variables constrained by it), normalised by the
number of constraints.

\item[Multiple shared variables] The proportion of pairs of constraints that
share more than one variable.

\item[Normalised mean constraints per variable] For each variable, we count
the number of constraints on the variable. The mean average is taken, and this
is normalised by dividing by the number of constraints.

\item[Ratio of auxiliary variables to other variables] Auxiliary variables
are introduced by decomposition of expressions in order to be able to express
them in the language of the solver. We use the ratio of auxiliary variables to
other variables.

\item[Tightness] The tightness of a constraint is the proportion of disallowed
tuples. The tightness is estimated by sampling 1000 random tuples (that are
valid w.r.t.\ variable domains) and testing if the tuple satisfies the
constraint. The tightness quartiles and the mean tightness over all constraints
is used.

\item[Proportion of symmetric variables] In many CSPs, the variables form
equivalence classes where the number and type of constraints a variable is in
are the same. For example in the CSP
\(x_1 \times x_2 = x_3, x_4 \times x_5 = x_6\),
\(x_1, x_2, x_4, x_5\) are all indistinguishable, as are \(x_3\) and \(x_6\).
The first stage of the algorithm used by Nauty~\cite{McKay81} detects this
property. Given a partition of \(n\) variables generated by this algorithm,
we transform this into a number between 0 and 1 by taking the proportion of
all pairs of variables which are in the same part of the partition.

\item[Alldifferent statistics] The size of the union of all variable domains in
an alldifferent constraint divided by the number of variables. This is a measure
of how many assignments to all variables that satisfy the constraint there are.
We used the quartiles and the mean over all alldifferent constraints.
\end{description}

In creating this set of attributes, we intended to cover a wide range of
possible factors that affect the performance of different alldifferent
implementations. Wherever possible, we normalised attributes that would be
specific to problem instances of a particular size.  This is based on the
intuition that similar instances of different sizes are likely to behave
similarly. Computing the features took 27 seconds per instance on average.

\section{Learning a problem classifier}

Before we used machine learning on the set of training instances, we annotated
each problem instance with the alldifferent implementation that had the best
performance on it according to the following criteria. If the na\"ive
alldifferent implementation took less CPU time than all the other ones, it was
chosen, else the implementation which had the best performance in terms of
search nodes per second was chosen. All implementations except the na\"ive one
explore the same search space.  If no solver was able to solve the instance, we
assigned a ``don't know'' annotation.

We used the WEKA~\cite{weka} machine learning software through the R~\cite{R}
interface to learn classifiers. We used almost all of the WEKA classifiers that
were applicable to our problem -- algorithms which generate decision rules,
decision trees, Bayesian classifiers, nearest neighbour and
neural networks. Our selection is broad and includes most major machine learning
methodologies. The specific classifiers we used are
\texttt{BayesNet},
\texttt{BFTree},
\texttt{ConjunctiveRule},
\texttt{DecisionTable},
\texttt{FT},
\texttt{HyperPipes},
\texttt{IBk},
\texttt{J48},
\texttt{J48graft},
\texttt{JRip},
\texttt{LADTree},
\texttt{MultilayerPerceptron},
\texttt{NBTree},
\texttt{OneR},
\texttt{PART},
\texttt{RandomForest},
\texttt{RandomTree},
\texttt{REPTree} and
\texttt{ZeroR},
all of which are described in~\cite{wekabook}.

For all of these algorithms, we used the default parameters provided by WEKA.
While the performance would have been improved by carefully tuning those
parameters, a lot of effort and knowledge is required to do so. Instead, we
used the standard parameter configuration which is applicable for other machine
learning problems as well and not specific to this paper.

The problem of classifying problem instances here is different to normal machine
learning classification problems. We do not particularly care about classifying
as many instances as possible correctly; we rather care that the instances that
are important to us are classified correctly. The higher the potential gain is
for an instance, the more important it is to us. If, for example, the difference
between making the right and the wrong decision means a difference in CPU time
of 1\%, we do not care whether the instance is classified correctly or not. If
the difference is several orders of magnitude on the other hand, we really do
want this instance to be classified correctly.

Based on this observation, we decided to measure the performance of the learned
classifiers not in terms of the usual machine learning performance measures, but
in terms of misclassification penalty~\cite{DBLP:journals/jair/XuHHL08}. The
misclassification penalty is the additional CPU time we require to solve a
problem instance when choosing to solve it with a solver that is not the fastest
one. If the selected solver was not able to solve the problem, we assumed the
timeout of 3600 seconds minus the CPU time the fastest solver took to be the
misclassification penalty.  This only gives the lower bound, but the correct
value cannot be estimated easily.

We furthermore decided to assign the maximum misclassification penalty (or the
maximum possible gain), cf.\ Figure~\ref{fig:speedup-potential}
as a cost to each instance as follows. To bias the WEKA classifiers towards the
instances we care about most, we used the common technique of duplicating
instances~\cite{wekabook}.  Each instance appeared in the new data set
$1+\left\lceil\log_2(\mathtt{cost})\right\rceil$ times. The particular formula
to determine how often each instance occurs was chosen empirically such that
instances with a low cost are not disregarded completely, but instances with a
high cost are much more important. Each instance will be in the data set used
for training the machine learning classifiers at least once and at most 13 times
for a theoretic maximum cost of 3600.

To achieve multi-level classification, each individual classifier below consists
of a combination of classifiers. First we make the decision whether to use
the alldifferent version equivalent to the binary decomposition or the other
one, then, based on the previous decision, we decide which specific version of
the alldifferent constraint to use.

Table~\ref{tab:cost} shows the total misclassification penalty for all
classifiers with and without instance duplication on the first data set. It
clearly shows that our cost model improves the performance significantly in
terms of misclassification penalty for almost all classifiers.

\begin{table}
\begin{center}
\begin{tabular*}{\textwidth}{c@{\extracolsep{\fill}}c}
\begin{tabular*}{.46\textwidth}{lr@{\extracolsep{\fill}}r}
 & \multicolumn{2}{c}{misclass. penalty [s]}\\
classifier & all equal & cost model\\
\hline
BayesNet & 1494 & 3.9\\
BFTree & 8.4 & 1.1\\
ConjunctiveRule & 2300 & 1433\\
DecisionTable & 249 & 1.6\\
FT & 248 & 1.2\\
HyperPipes & 867 & 867\\
IBk & 109 & 109\\
J48 & 8.2 & 1.2\\
J48graft & 8.2 & 1.2\\
JRip & 283 & 1.3\\
\end{tabular*}
&
\begin{tabular*}{.51\textwidth}{lr@{\extracolsep{\fill}}r}
 & \multicolumn{2}{c}{misclass. penalty [s]}\\
classifier & all equal & cost model\\
\hline
LADTree & 8.4 & 6.5\\
MultilayerPerceptron & 249 & 8.5\\
NBTree & 9 & 1.3\\
OneR & 69.5 & 409\\
PART & 5.9 & 1\\
RandomForest & 41.9 & 0.9\\
RandomTree & 1 & 1\\
REPTree & 1099 & 10.8\\
ZeroR & 2304 & 2304\\
\\
\end{tabular*}
\end{tabular*}
\smallskip
\caption{Misclassification penalty for all classifiers with and without
instances duplicated according to their cost in the training data set. All
numbers are rounded.}
\label{tab:cost}
\end{center}
\end{table}

For each classifier, we did stratified $n$-fold cross-validation -- the original
data set is split into $n$ parts of roughly equal size. Each of the $n$
partitions is in turn used for testing. The remaining $n-1$ partitions are used
for training. In the end, every instance will have been used for both training
and testing in different runs~\cite{wekabook}. Stratified cross-validation
ensures that the ratio of the different classification categories in each subset
is roughly equal to the ratio in the whole set. If, for example, about 50\% of
all problem instances in the whole data are solved fastest with the na\"ive
implementation, it will be about 50\% of the instances in each subset as well.

There are several problems we faced when generating the classifiers. First, we
do not know which one of the machine learning algorithms was suited best for
our classification problem; indeed we do not know whether the features of the
problem instances we measured are able to capture the factors which affect the
performance of each individual implementation at all. Second, the learned
classifiers could be overfitted. We could evaluate the performance of each
classifier on the second set of problem instances and compare it to the
performance during machine learning to assess whether it might be overfitted.
Even if we were able to reliably detect overfitting this way, it is not obvious
how we would change or retrain the classifier to remove the overfitting.
Instead, we decided to use all classifiers -- for each machine learning
algorithm the $n$ different classifiers created during the $n$-fold
cross-validation and the classifiers created by each different machine learning
algorithm.

We decided to use three-fold cross-validation as an acceptable compromise
between trying to avoid overfitting and time required to compute and run the
classifiers. We combine the decisions of the individual classifiers by majority
vote. The technique of combining the decisions of several classifiers was
introduced in~\cite{adaboost} and formalised in~\cite{ensemble}.

Table~\ref{tab:final} shows the overall performance of our meta-classifier
compared to the best and worst individual classifier for each set and several
other hypothetical classifiers. Our meta-classifier outperforms a classifier
which always makes the default decision even on the second set of problem
instances.  This set is an extreme case because just making the default choice
is almost always the best choice -- the misclassification penalty for the
default choice classifier is extremely low given the large number of instances.
Even though there is only very little room for improvement (cf.\
Figure~\ref{fig:speedup-potential}), we achieve some of it.

\begin{table}
\begin{center}
\begin{tabular*}{\textwidth}{l@{\extracolsep{\fill}}rr@{\extracolsep{\fill}}rr}
 & \multicolumn{4}{c}{misclassification penalty [s]}\\
 & \multicolumn{2}{c}{instance set 1} & \multicolumn{2}{c}{instance set 2}\\
classifier & all features & cheap features & all features & cheap features\\
\hline
oracle & 0 & 0 & 0 & 0\\
anti-oracle & 19993 & 19993 & 47144 & 47144\\
\textbf{default decision} & \textbf{2304} & \textbf{2304} & \textbf{223} & \textbf{223}\\
random decision & 5550 & 5550 & 564 & 564\\
best classifier on set 1 & 0.998 & 0.994 & 131 & 220.3\\
worst classifier on set 1 & 2304 & 2304 & 223 & 223\\
best classifier on set 2 & 0.998 & 61.66 & 131 & 186\\
worst classifier on set 2 & 1.34 & 1.44 & 621 & 610 \\
\textbf{meta-classifier} & \textbf{1.16} & \textbf{0.996} & \textbf{220} &
\textbf{222.95}
\end{tabular*}
\smallskip
\caption{Summary of classifier performance on both sets of benchmarks in terms
of total misclassification penalty in seconds. We first evaluated the
performance using the full set of features described in Section~\ref{sec:attrs},
then using only the cheap features. The oracle classifier always makes the right
decision, the anti-oracle always the worst possible wrong decision. The
``default decision'' classifier always makes the same decision and the ``random
decision'' one chooses one of the possibilities at random. Three-fold
cross-validation was used. All numbers are rounded.}
\label{tab:final}
\end{center}
\end{table}

It also shows that the classifiers we have learned on a data set that contains
problem instances from many problem classes can be applied to a different data
set with instances from different problem classes and still achieve a
performance improvement. Based on this observation, we suggest that
our meta-classifier is generally applicable.

Another observation we made is that the performance of the meta-classifier does
not suffer even if a large number of the classifiers that it combines perform
badly individually. This suggests that the classifiers complement each other --
the set of instances that each one misclassifies are different for each
classifier. Note also that the classifier which performs best on one set of
instances is not necessarily the best performer on the other set of instances.
The same observation can be made for the classifier with the worst performance
on one of the instance sets. This means that we cannot simply choose ``the
best'' classifier or discard ``the worst'' for a given set of training
instances.  Table~\ref{tab:bestworst} provides further evidence for this. The
individual best and worst classifiers vary not only with the data set, but also
with the set of features used.

\begin{table}
\begin{center}
\begin{tabular*}{\textwidth}{l@{\extracolsep{\fill}}rr@{\extracolsep{\fill}}rr}
& \multicolumn{2}{c}{instance set 1} & \multicolumn{2}{c}{instance set 2}\\
& all features & cheap features & all features & cheap features\\
\hline
best classifier & \texttt{IBk} & \texttt{BFTree} & \texttt{IBk} & \texttt{BayesNet}\\
worst classifier & \texttt{ZeroR} & \texttt{ZeroR} & \texttt{LADTree} & \texttt{LADTree}
\end{tabular*}
\smallskip
\caption{Individual best and worst classifiers for the different data and
feature sets for the numbers presented in Table~\ref{tab:final}.}
\label{tab:bestworst}
\end{center}
\end{table}

The time required to compute the features was 27 seconds per instance on
average, and it took 0.2 seconds per instance on average to run the classifiers
and combine their decisions.  If we take this time into account, our system is
slower than just using the default implementation. This is mostly because of the
cost of computing all the features required to make the decision. We do however
learn good classifiers in the sense that the decision they make is better than
just using the standard implementation.

We now focus on making a decision as quickly as possible. Most of the time
required to make the decision is spent computing the features that the
classifiers need. We removed the most expensive features -- all the properties
of the primal graph described in Section~\ref{sec:attrs} apart from edge
density.

The results for the reduced set of features are shown in Table~\ref{tab:final}
as well. The performance is not significantly worse and even better on the first
set of instances, but the time required to compute all the features is only
about 3 seconds per instance. On the first set of benchmarks, we solve each
instance on average 8 seconds faster using our system (misclassification penalty
of default decision minus that of our system divided by the number of instances
in the set). We are therefore left with a performance improvement of an average
of 5 seconds per instance. On the second set, we cannot reasonably expect a
performance improvement -- the perfect oracle classifier only achieves about 0.2
seconds per instance on average.

\begin{figure}
\begin{center}
\includegraphics[width=.8\textwidth]{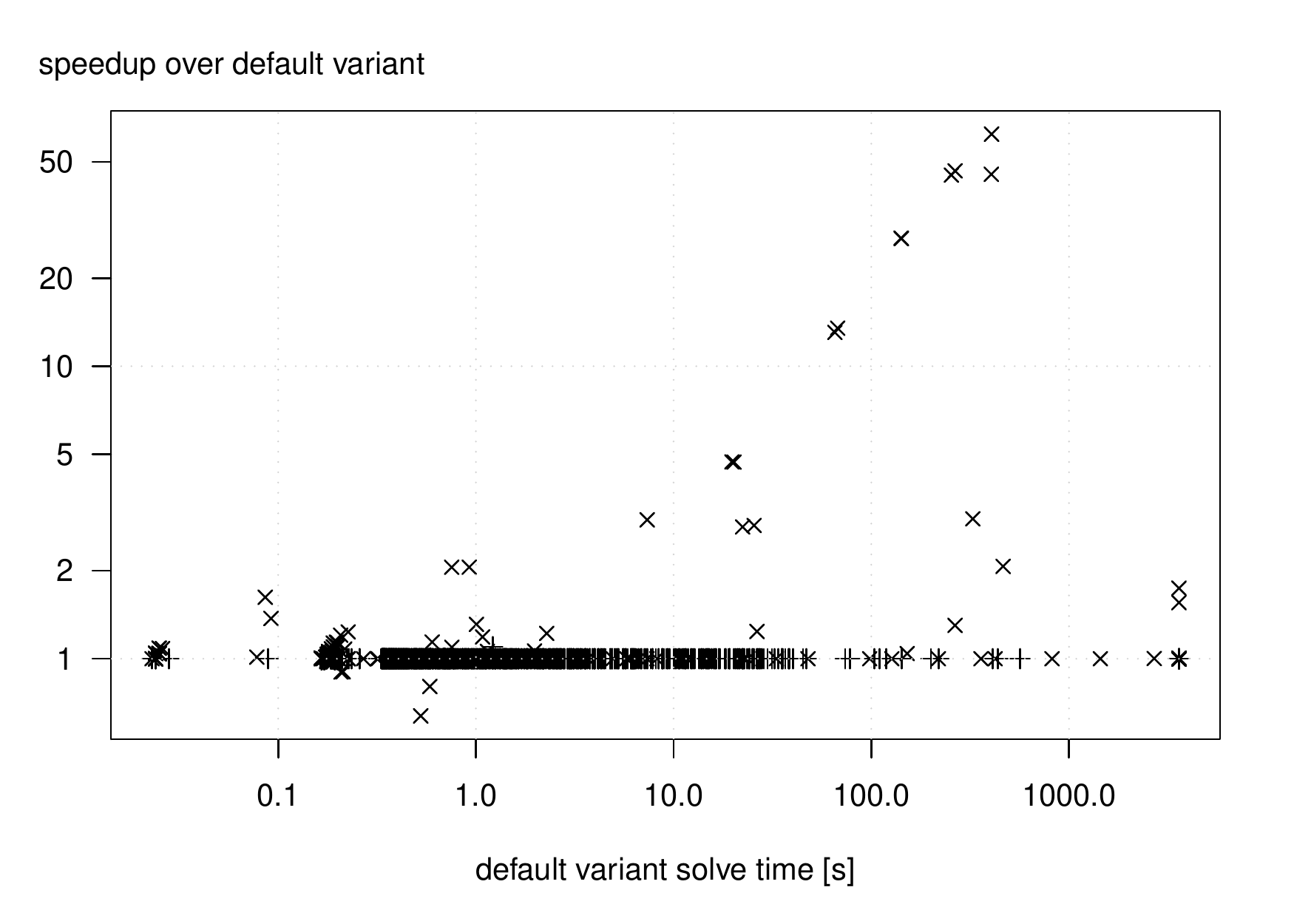}
\end{center}
\vspace*{-3em}
\caption{Speedup achieved by the meta-classifier using the set of
cheaply-computable features. The figure does not take the overhead of computing
the features and running the classifiers into account. The crosses represent the
instances of the first data set, the pluses the instances of the second data
set.}
\label{fig:speedup-achieved}
\end{figure}

Figure~\ref{fig:speedup-achieved} revisits Figure~\ref{fig:speedup-potential}
and shows the actual speedup our meta-classifier achieves for each instance. It
convincingly illustrates the quality of our classifier. The instances where we
suffer a slowdown are ones that are solved almost instantaneously, whereas the
correctly classified instances are the hard ones that we care about most. In
particular the instances where a large speedup can be gained are classified
correctly by our system.

\section{Conclusions and future work}

We have applied machine learning to a complex decision problem in constraint
programming. To facilitate this, we evaluated the performance of constraint
solvers representing all the decisions on two large sets of problem instances.
We have demonstrated that training a set of classifiers without intrinsic
knowledge about each individual one and combining their decisions can improve
performance significantly over always making a default decision. In particular,
our combined classifier is almost as good as the best classifier in the set and
much better than the worst classifier while mitigating the need to select and
tune an individual classifier.

We have conclusively shown that we can improve significantly on default
decisions suggested in the state-of-the-art literature using a relatively simple
and generic procedure. We provide strong evidence for the general applicability
of a set of classifiers learned on a training set to sets of new, unknown
instances. We identified several problems with using machine learning to make
constraint programming decisions and successfully solved them.

Our system achieves performance improvements even taking the time it takes to
compute the features and run the learned classifiers into account. For atypical
sets of benchmarks, where always making the default decision is the right choice
in almost all of the cases, we are not able to compensate for this overhead, but
we are confident that we can achieve a real speedup on average.

We have identified two major directions for future research. First, it would be
beneficial to analyse the individual machine learning algorithms and evaluate
their suitability for our decision problem. This would enable us to make a more
informed decision about which ones to use for our purposes and may suggest
opportunities for improving them.

Second, selecting which features of problem instances to compute is a
non-trivial choice because of the different cost and benefit associated with
each one. The classifiers we learned on the reduced set of features did not seem
to suffer significantly in terms of performance. Being able to assess the
benefit of each individual feature towards a classifier and contrast that to the
cost of computing it would enable us to make decisions of equal quality cheaper.

\section*{Acknowledgements}

The authors thank Chris Jef\/ferson for providing some of the feature
descriptions. We thank Jesse Hoey for useful discussions about machine learning
and the anonymous reviewers for their feedback.  Peter Nightingale is supported
by EPSRC grants EP/H004092/1 and EP/E030394/1.  Lars Kotthof\/f is supported by a SICSA
studentship.

\bibliography{alldiff}

\begin{thebibliography}{10}
\providecommand{\url}[1]{\texttt{#1}}
\providecommand{\urlprefix}{URL }

\bibitem{DBLP:conf/cp/AnsoteguiST09}
Ans{\'o}tegui, C., Sellmann, M., Tierney, K.: A gender-based genetic algorithm
  for the automatic configuration of algorithms. In: CP. pp. 142--157 (2009)

\bibitem{DBLP:conf/ecai/BorrettTW96}
Borrett, J., Tsang, E., Walsh, N.: Adaptive constraint satisfaction: The
  quickest first principle. In: ECAI. pp. 160--164 (1996)

\bibitem{constraint-processing-dechter}
Dechter, R.: Constraint Processing. Elsevier Science (2003)

\bibitem{ensemble}
Dietterich, T.G.: Ensemble methods in machine learning. In: First International
  Workshop on Multiple Classifier Systems. pp. 1--15 (2000)

\bibitem{DBLP:conf/cp/EpsteinFWMS02}
Epstein, S., Freuder, E., Wallace, R., Morozov, A., Samuels, B.: The adaptive
  constraint engine. In: CP. pp. 525--542 (2002)

\bibitem{adaboost}
Freund, Y., Schapire, R.E.: A decision-theoretic generalization of on-line
  learning and an application to boosting. In: {EuroCOLT}. pp. 23--37 (1995)

\bibitem{lazyecai}
Gent, I., Jefferson, C., Kotthoff, L., Miguel, I., Moore, N., Nightingale, P.,
  Petrie, K.: Learning when to use lazy learning in constraint solving. In:
  ECAI (2010)

\bibitem{minion}
Gent, I., Jefferson, C., Miguel, I.: Minion: A fast scalable constraint solver.
  In: ECAI. pp. 98--102 (2006)

\bibitem{lazylearning}
Gent, I., Miguel, I., Moore, N.: Lazy explanations for constraint propagator.
  In: PADL (2010)

\bibitem{petealldiff}
Gent, I., Miguel, I., Nightingale, P.: Generalised arc consistency for the
  alldifferent constraint: An empirical survey. Artif. Intell.  172(18),
  1973--2000 (2008)

\bibitem{guerri-milano-model-selection}
Guerri, A., Milano, M.: Learning techniques for automatic algorithm portfolio
  selection. In: ECAI. pp. 475--479 (2004)

\bibitem{weka}
Hall, M., Frank, E., Holmes, G., Pfahringer, B., Reutemann, P., Witten, I.: The
  {WEKA} data mining software: An update. SIGKDD Explorations  11(1) (2009)

\bibitem{hoeve_alldifferent_2001}
van Hoeve, W.J.: The alldifferent Constraint: A Survey (2001)

\bibitem{DBLP:conf/cp/HutterHHL06}
Hutter, F., Hamadi, Y., Hoos, H., Leyton-Brown, K.: Performance prediction and
  automated tuning of randomized and parametric algorithms. In: CP. pp.
  213--228 (2006)

\bibitem{DBLP:conf/ijcai/KhudaBukhshXHL09}
KhudaBukhsh, A., Xu, L., Hoos, H., Leyton-Brown, K.: {SAT}enstein:
  Automatically building local search {SAT} solvers from components. In: IJCAI.
  pp. 517--524 (2009)

\bibitem{survey}
Kotthof\/f, L.: Constraint solvers: An empirical evaluation of design
  decisions. CIRCA preprint (2009),
  \url{http://www-circa.mcs.st-and.ac.uk/Preprints/solver-design.pdf}

\bibitem{lagoudakis}
Lagoudakis, M., Littman, M.: Reinforcement learning for algorithm selection.
  In: AAAI/IAAI. p. 1081 (2000)

\bibitem{xcsprepo}
Lecoutre, C.: {XCSP} benchmarks. \url{http://tinyurl.com/y6hpphs} (June 2010)

\bibitem{nudelman}
Leyton-Brown, K., Nudelman, E., Andrew, G., McFadden, J., Shoham, Y.: A
  portfolio approach to algorithm selection. In: IJCAI. pp. 1542--1543 (2003)

\bibitem{McKay81}
McKay, B.: Practical graph isomorphism. In: Numerical mathematics and
  computing, Proc.~10th Manitoba Conf., Winnipeg/Manitoba 1980,
  Congr.~Numerantium 30. pp. 45--87 (1981), see also
  \url{http://cs.anu.edu.au/people/bdm/nauty}

\bibitem{DBLP:journals/constraints/Minton96}
Minton, S.: Automatically configuring constraint satisfaction programs: A case
  study. Constraints  1(1/2),  7--43 (1996)

\bibitem{cphydra}
O'Mahony, E., Hebrard, E., Holland, A., Nugent, C., O'Sullivan, B.: Using
  case-based reasoning in an algorithm portfolio for constraint solving. In:
  19th Irish Conference on AI (2008)

\bibitem{modelrun}
Puget, J.F.: Constraint programming next challenge: Simplicity of use. In: CP.
  pp. 5--8 (2004)

\bibitem{R}
{R Development Core Team}: R: A Language and Environment for Statistical
  Computing. R Foundation for Statistical Computing (2009)

\bibitem{reginalldiff}
R\'{e}gin, J.C.: A filtering algorithm for constraints of difference in {CSPs}.
  In: AAAI. pp. 362--367 (1994)

\bibitem{rice}
Rice, J.: The algorithm selection problem. Advances in Computers  15,  65--118
  (1976)

\bibitem{watts-strogatz-small-world}
Watts, D., Strogatz, S.: Collective dynamics of `small-world' networks. Nature
  393,  440--442 (1998)

\bibitem{wekabook}
Witten, I., Frank, E.: Data Mining: Practical Machine Learning Tools and
  Techniques with Java Implementations. Morgan Kaufmann (2005)

\bibitem{DBLP:journals/jair/XuHHL08}
Xu, L., Hutter, F., Hoos, H., Leyton-Brown, K.: {SAT}zilla: Portfolio-based
  algorithm selection for {SAT}. J. Artif. Intell. Res. (JAIR)  32,  565--606
  (2008)

\end{thebibliography}
\bibliographystyle{splncs03}

\end{document}